\documentclass[10pt,twocolumn,letterpaper]{article}

\usepackage{cvpr}
\usepackage{times}
\usepackage{epsfig}
\usepackage{graphicx}
\usepackage{amsmath}
\usepackage{amssymb}
\usepackage{textcomp}
\usepackage{subcaption}
\usepackage{siunitx}
\usepackage{booktabs}
\usepackage[table]{xcolor}

\makeatletter
\@namedef{ver@everyshi.sty}{}
\makeatother
\usepackage{pgfplotstable}

\pgfplotsset{every axis plot post/.append style={solid,line width=1pt}}

\usepackage[pagebackref=true,breaklinks=true,letterpaper=true,colorlinks,bookmarks=false]{hyperref}

\cvprfinalcopy 

\ifcvprfinal\fi
\begin{document}

\title{Congestion Analysis of Convolutional Neural Network-Based Pedestrian Counting Methods on Helicopter Footage}

\author{
\begin{minipage}{.25\textwidth}\center{Gergely Cs\"onde\\
The University of Tokyo, Department of Civil Engineering\\
4-6-1 Komaba, Meguro, Tokyo 1538505, JAPAN\\
{\tt\small csonde@iis.u-tokyo.ac.jp}}
\end{minipage}
\and
\begin{minipage}{.25\textwidth}\center{Yoshihide Sekimoto\\
The University of Tokyo, Institute of Industrial Science\\
4-6-1 Komaba, Meguro, Tokyo 1538505, JAPAN\\
{\tt\small sekimoto@iis.u-tokyo.ac.jp}}
\end{minipage}
\and
\begin{minipage}{.25\textwidth}\center{Takehiro Kashiyama\\
The University of Tokyo, Institute of Industrial Science\\
4-6-1 Komaba, Meguro, Tokyo 1538505, JAPAN\\
{\tt\small ksym@iis.u-tokyo.ac.jp}}
\end{minipage}
}

\maketitle

\begin{abstract}
   Over the past few years, researchers have presented many different applications for convolutional neural networks, including those for the detection and recognition of objects from images. The desire to understand our own nature has always been an important motivation for research. Thus, the visual recognition of humans is among the most important issues facing machine learning today. Most solutions for this task have been developed and tested by using several publicly available datasets. These datasets typically contain images taken from street-level closed-circuit television cameras offering a low-angle view. There are major differences between such images and those taken from the sky. In addition, aerial images are often very congested, containing hundreds of targets. These factors may have significant impact on the quality of the results. In this paper, we investigate state-of-the-art methods for counting pedestrians and the related performance of aerial footage. Furthermore, we analyze this performance with respect to the congestion levels of the images.
\end{abstract}

\section{Introduction}
The resurfacing and growing development of convolutional neural networks (CNN) has brought many problems into focus, which would have been too difficult to solve in the past with classic image-processing techniques. The recognition of humanoid shapes is a problem that intuitively comes up in daily life.

There are several potential usages of CNN-based pedestrian analysis. We are interested in reconstructing the flow of people in a given urban area. Data models supporting this pursuit are used for many different purposes. For example, they are used as inputs for designing certain infrastructure objects or estimating traffic changes caused by temporal obstructions, such as constructions or disasters.

These methods have come a long way since CNNs matured in the early 2010s. The main focus of many studies has been the problem of different object scales. Currently, the best methods can identify objects of the same class at different scales without significant overhead. Therefore, we introduce a new type of footage that provides a challenge for models trained on conventional datasets. This footage is created at high altitudes with steep angles over dense urban areas. With this footage, the range of scales are small because of the geometry of the setup. However, the range of various congestion levels is large, which can significantly affect the accuracy. A group of loosely scattered people is easier to count, whereas a tight clump is often better for estimation. This could be considered an analogy for object detection and regression/density-map methods in general. Therefore, we suggest that the different methods will yield good results for different congestion levels. Our initial plan was to design and test a network that combines other models with strengths at different congestion levels, switching between them accordingly. Thus, we formalized what we meant by congestion and then checked if it proved our assumption.

In this paper, we investigate how congestion affects different crowd-counting methods by surveying the accuracy of state-of-the-art detection-based and density map-based methods, which are either general purpose or specifically developed for human-shape recognition and counting. Furthermore, we introduce two metrics for measuring congestion levels originating from crowd density. Then, we take our newly introduced aerial footage and divide it into several subsets of different congestion levels. On these subsets, we test the selected methods and analyze their accuracy on the levels of congestion. Our results suggest that the initially proposed combination model design does not offer any benefit. Therefore, we do not follow up on that pursuit. Instead we show that a general-purpose method can be improved for a specific task by optimizing its loss function.

\section{Previous work}

\subsection{Existing models}
In this paper, we focus on static situations. The task of people-counting is fundamentally divided into two types: direct and indirect methods. The former involves detecting each individual target. The result can be obtained by counting people one-by-one. In this case, geometric information, such as central coordinates or bounding boxes, are acquired for every detected person. For the indirect method, instead of the count or the individual geometric data, more abstract data are produced. Then the count is derived. There is a third approach (i.e., simple regression) in which there is only one output: the estimated number. However, thus far, this method has not been competitive.

One of the most accurate object detection networks is Faster R-CNN \cite{Ren:2015:FRT:2969239.2969250}, which was developed from methods using classical image processing steps (\cite{Girshick:2015:FR:2919332.2920125}, \cite{Girshick2014RichFH}). The model has a feature extractor backbone, which is typically a popular classifier such as ResNet \cite{He2016DeepRL} or Inception \cite{Szegedy15Inception}. A region proposal network (RPN) extracts region proposals which are then propagated to a classifier network. The regions vary in size. In order to solve this issue spatial pyramid pooling (SPP) \cite{He14SPP} is used but only with one pyramid layer. A set number of regions can be proposed for every feature pixel which can reach a very high count. Classifying each and every one of these proposals forces the network to process the same area multiple times which is slow.

The other important attribute, apart from accuracy, is computational performance. Training and leveraging neural networks are computationally very expensive. Thus, any improvement that can reduce resource consumption is worth examining. Single-stage methods, which only pass through the image once, outperform Faster R-CNN in this process. One of the first representatives for this was the ``you only look once’’ (YOLO) architecture \cite{Redmon2016YOLO}. It has a very simple architecture. The validity of the method lies with how it grasps the task. Similar to Faster R-CNN, it predicts a set number of bounding boxes for every cell in the deepest feature map, but it does not use anchor boxes. Predictions are attached to the grid cell containing the center of the bounding box. Faster R-CNN uses a different network for classification, and YOLO uses the same. 

Single-shot object detectors (SSD) \cite{Liu2016SSDSS} tried to overcome YOLO’s shortcomings. The base was VGG-16. The problem of different scales was tackled by making predictions from several scale-feature maps.

YOLOv2 truly outperformed YOLO in its accuracy and speed \cite{Redmon2017YOLO9000BF}. Several improvements were applied, including batch normalization, a fully convolutional architecture, and ``bounding-box priors,’’. A unique approach included the ``fine-grained features.’’

With YOLOv3 \cite{DBLP:journals/corr/abs-1804-02767}, there were only minor improvements. The improved feature extractor is worth mentioning, and the classification layers were altered. Classification was done on three different scales, each with its own bounding-box priors. However, the channels for each scale came from multiple scales by up-sampling lower resolution feature maps. This solution is similar to the feature pyramid networks (FPN) \cite{Lin17FPN}.

Regarding indirect methods, we investigated density map-based solutions. The premise of this solution is that the output of the network is an image or pixel map rather than a bounding-box list. Each output pixel has a single value indicating the average number of people in that pixel. Naturally, this value is most likely a fraction. The count can be achieved by integrating over any given area.

There are several difficulties with this method. Many teams have attempted to address these problems in a variety of ways. In fact, a survey was performed on such methods \cite{Sindagi17Surv}. For more recent information, one should visit the GitHub repository called ``Awesome Crowd Counting’’ \cite{AwesomeCC}. As this paper was being written, the repository contained the most recent and most relevant papers about density map-based pedestrian counting solutions. Accuracy metrics are listed on the commonly spread datasets, and source codes are shared.

We only consider solutions having available online implementations for our experiments. One such method is the multi-column CNN (M-CNN) \cite{Zhang_2016_CVPR}. The structure is very simple and uses three parallel CNNs of the same type. The only differences lie with the size of kernels and the number of filters. Each network learns people at different scales. The ends of the output channels of each network are concatenated to the rest, and predictions are made using a $1\times1$ convolution. The output resolution is $1/4$ of the original.

Many networks (e.g., VGG-16) scale down their output by large factors (e.g., 32 in the case of VGG-16) because of their pooling layers. The Congested Scene Recognition Network (CSRNet) \cite{Li_2018_CVPR} attempted to solve this problem using dilated convolutions. In CSRNet, a VGG-16 base is used. However, instead of five pooling layers, there are only three. To replace the missing large-scale features, dilated convolutions are used. The final output resolution is $1/8$ of the original.

The Context-Aware Network (CAN) \cite{DBLP:journals/corr/abs-1811-10452} uses rather complicated network architecture. Simply combining results from a set of different scales does not yield better result for people at intermediate scales. Thus, what CAN achieves is successfully using an SPP to obtain different scale feature maps from a VGG-16 base. Instead of simply concatenating these feature maps, it applies weights to them. These weights are also learned by the network.

\begin{figure}[t!]
  \begin{center}
    \framebox[\linewidth]{\includegraphics[width=\linewidth]{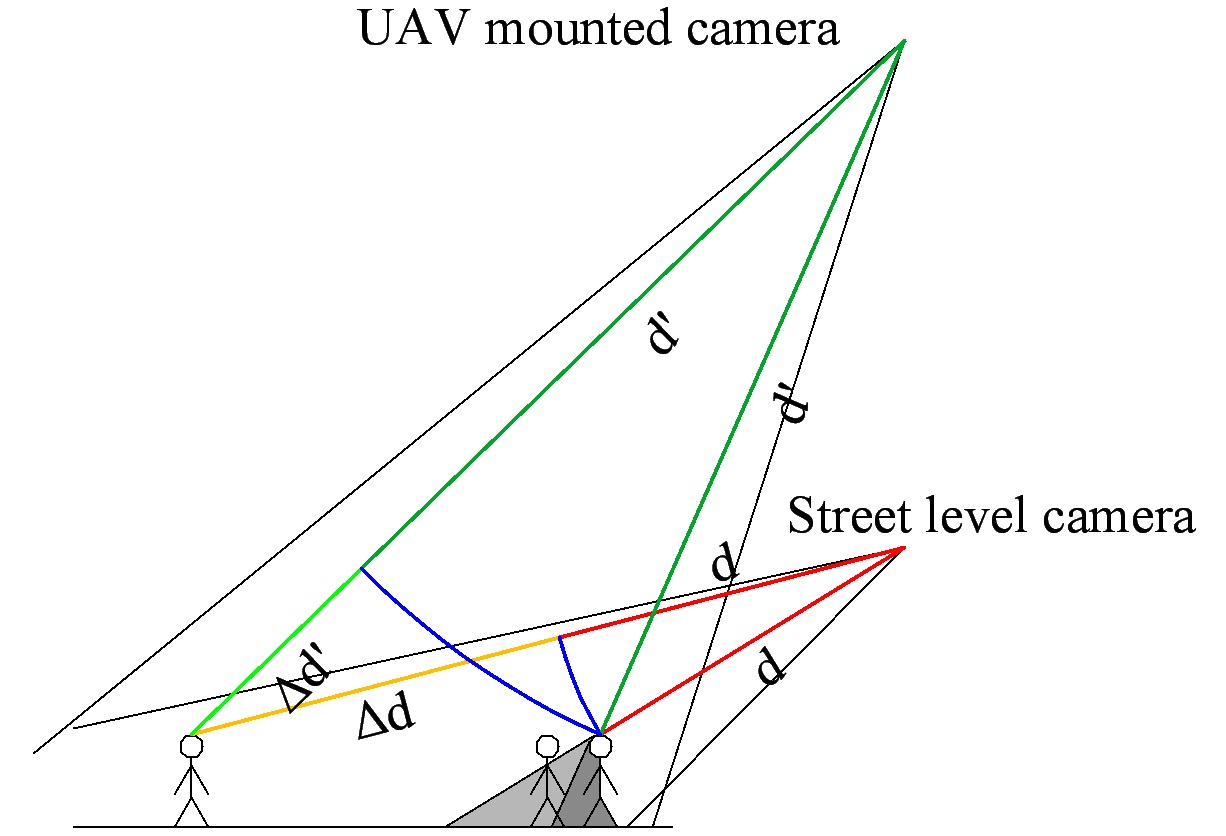}}
    \caption{Illustration of the differences between street-level and aerial footage. $(d+{\Delta}d)/d$ is much farther from 1.0 than $(d’+{\Delta}d’)/d’$. Additionally, the dark gray area is obstructed from both cameras, whereas the light gray area is only obstructed from the street-level camera. Note that the image is not to scale.}
    \label{fig:aerialf}
  \end{center}
\end{figure}

\subsection{Existing Datasets}

Many CNN-based detecting or counting methods have been developed and tested on standard datasets, such as UCSD \cite{Chan08UCSD}, INRIA \cite{Dalal:2005:HOG:1068507.1069007}, Caltech Benchmark \cite{Dollar09pedestriandetection:}, UCF\_CC\_50 \cite{Idrees:2013:MMC:2514950.2515922}, and ShanghaiTech \cite{Zhang_2016_CVPR}. Although these datasets can be distinguished from each other by the average person count, they have one thing in common: they are taken from near-street-level cameras at low angles. Therefore, these images contain many obstructions and perspective distortions. Both issues pose much less difficulty for aerial footage.

Obstructions are caused when people walk next to each other. On images taken from a steep angle, the obstruction is not very big, because people normally do not walk over top of each other. Perspective distortion is caused by the relative difference between the distances of objects from the camera. The size of an object on an image is inversely proportional to its distance from the camera. Thus, if the ratio of two objects’ distances is much larger (or smaller) than 1.0, the difference in their size on the image will also be very big. On photos taken from a high altitude, this ratio stays close to 1.0. Figure \ref{fig:aerialf} illustrates these issues.

Aerial datasets are most recently created using airplane or drone platforms. Airplane footage is typically taken from higher altitudes, which is not suitable for the detection of small targets. In the case of drone datasets, elevations are quite a bit lower. Thus, human detection can be achieved. Often, the angle is vertical, which is advantageous when considering occlusion. However, human features are less recognizable. If the angle is low, then perspective distortions occur. However, it provides a better viewpoint for human features. Examples for drone datasets include SDDs \cite{Stanforddrone}, VisDrone2019\cite{zhuvisdrone2018}, and Okutama-Action \cite{Barekatain_2017_CVPR_Workshops}.

Some of these datasets are image based, whereas others are video-based. The image-based datasets were annotated manually. The video-based datasets share a common annotation strategy of annotating a few key frames (e.g., every 10th). Then, it interpolates the annotation between frames. This vastly increases the size of the dataset, but not the unique object instance count. Some video-based datasets (i.e., SDD and Okutama-Action) use a unique identification-based annotation. Thus, it can be also be used for tracking. Furthermore there are occlusion and truncation information available in a few datasets.

\begin{figure}[h]
  \begin{center}
    \includegraphics[width=\linewidth]{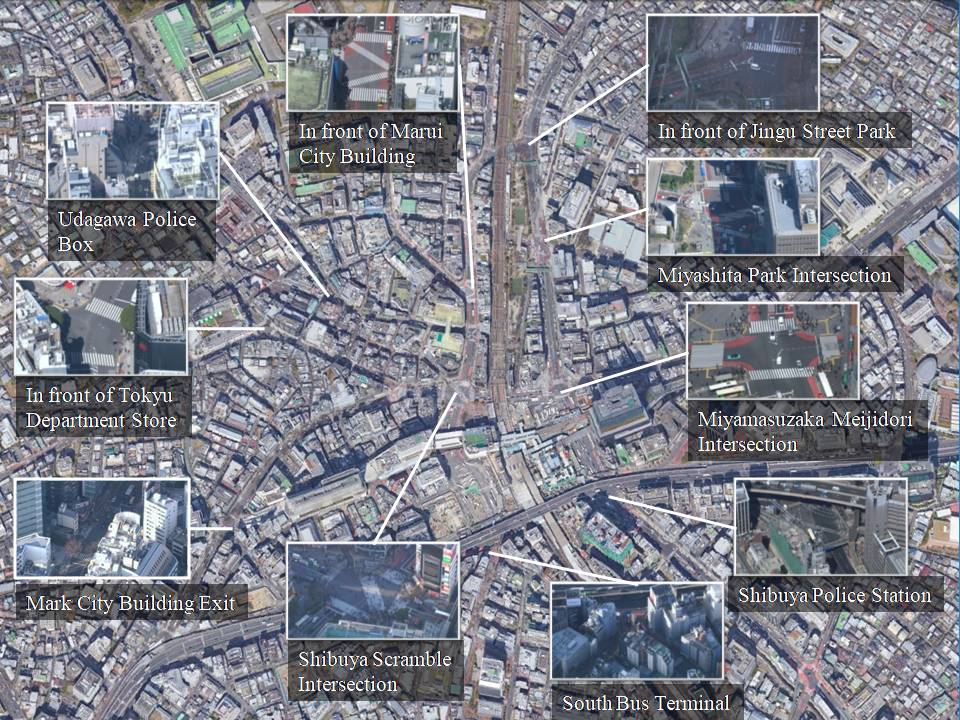}
    \caption{Locations of the intersections where the source footage for the TH2019 dataset was created.}
    \label{fig:footageloc}
  \end{center}
\end{figure}

\section{Dataset}

We use a custom dataset for our experiments. It contains snapshots from full high-definition video footage taken from a helicopter over a dense urban area in Tokyo, Japan. It is named ``Tokyo Hawkweye 2019’’ or ``TH2019’’ for short. We are not aware of other datasets that cover the altitudes between that of drones and airplanes. All images were taken in the daytime under both sunny and cloudy weather conditions. The angle of the camera orientation relative to the horizontal plane was rather steep, greater than 45\textdegree, and the altitude was several-hundred meters. These facts present two main advantages: there are less obstructions; and the size of the targets on the image is relatively the same. These factors make the counting task easier. However, because of the high altitude and steep angles, the size of the targets is rather small, making the counting task more difficult.

VisDrone2019 and Okutama-Action, by comparison, both exhibit a similar point-of-view as our dataset. However, SDD uses a strictly vertical camera angle that makes human features more difficult to detect. All three datasets were from low altitudes ($<100m$) and exhibit strong perspective distortions. In our dataset, optical zoom further reduces perspective distortions, and the resulting images are near isometric. Of the mentioned datasets, only VisDrone2019 contained extremely crowded scenes. 

The average size of a target person in our footage was 10--25 pixels horizontally and 20--50 pixels vertically. If the original $1,920\times1080$ images were scaled down to about $600\times600$ (a rational size for training networks), the target size would change to 3--7 pixels by 11--27 pixels, which is very small. Thus, instead of scaling down, we split up the images into four smaller ($960\times540$) resolution images, and we used these as training data. These images still required scaling in most situations, but changes in the target size was not dramatic. Of course, splitting full HD images into four parts implies that detection is four-times slower for a whole image. However, if the results were a hundred times better, it is a good trade-off.

We obtained video footage from 10 different locations at three different periods. Figure \ref{fig:footageloc} shows the source locations. Not every combination was available. Thus, we settled for less than 30 videos. Every video is a few minutes long, which includes 2--3 traffic-lamp cycles. This is important, because, although we have hundreds of thousands of frames, many of them are essentially the same. When traffic lights are red, the same people stand still in the same location. If they are moving, they are still the same people, and, at this target size, the variation in their features is not very different than what can be achieved with data augmentation. Thus, our strategy with image sampling was to take 1--3 images per traffic-lamp cycle. The sample count is therefore biased towards less-crowded images, because we were aiming for an even pedestrian distribution among different crowd congestion levels. We will further elaborate this method in section \ref{Methodology}.

We created a dataset containing 296 images with 7,887 annotated pedestrians. For object detection-based and density map-based methods, the annotation format is different. For object detection-based solutions, we used bounding boxes. However, for density maps, we annotated the heads on which a Gaussian kernel was applied.
\section{Methodology}\label{Methodology}

\subsection{General accuracy analysis}
This work comprises two major parts. The first is a survey of the overall performance of the investigated CNNs on the complete TH2019 dataset. We split the dataset into training and test sets with a roughly 80--20\% annotation ratio. We trained one model for each network and evaluated them all. The best metrics choices are mean absolute error (MAE) and root mean-squared error (RMSE), because these can be evaluated over any method. For object detection-based methods, we also collect precision and recall at a 50\% confidence threshold with 0.5 intersection-over-union (IoU). The commonly used metric for object detectors is mean average precision (mAP). We will report these values in some tables.

\subsection{Congestion analysis}
The second major consideration is congestion analysis. We clarify the meaning of ``congestion’’ before analyzing it. For the introduction, we already mentioned that the definition task should be approached from the direction of crowd density.

Let us assume that we have two artificial images. Both are the same size containing exactly the same person at several different locations. In one image, we have 100 instances completely separated. In the other, we have 20 instances clumped together with many overlaps. A human could quickly count the instances one-by-one on the first image but may have difficulties separating the clumped targets on the second. Thus, in this situation, the image with fewer targets seems to be more congested. The overall density of the first image is higher, because there are more targets in the same area. However, the second image has a much higher local density in the clumped area. Although it is true that there is a correlation between count and congestion, the images having a higher person count are more likely to have clumps of people. Considering this, we follow two approaches in our analysis. One is based on density and the other is based on count.

\subsubsection{Density-based congestion}
In one approach, we use the result provided by the complete TH2019 dataset. We divide the test set into three congestion categories using k-means clustering. We define congestion for one image, as follows:
\begin{equation}
  \label{eq:1}
  C=\frac{\|\vec{d}\|_{L_{1}}}{\|\vec{d}\|_{L_{0}}},
\end{equation}
where $\vec{d}$ is the ground-truth density map of the image. What we need is count divided by occupied area. In the nominator 
we sum all the pixel values. That should equal the pedestrian count. In the denominator we would need to write the area of pedestrian masks but we do not have that. Instead we count the number of pixels having nonzero values on the ground-truth map. This definition can be calculated from the ground truth density map thus it directly reflects the input of the networks.

\begin{figure}[t!]
\centering
\begin{tikzpicture}
\begin{axis}[xlabel=count,ylabel=congestion]
	\addplot+[only marks]
	table [x index=0, y index=1, col sep=comma]{CSV/congcountcorr.csv};
\end{axis}
\end{tikzpicture}
\caption{Correlation between ground-truth count and congestion.}
\label{fig:countcongcorr}
\end{figure}
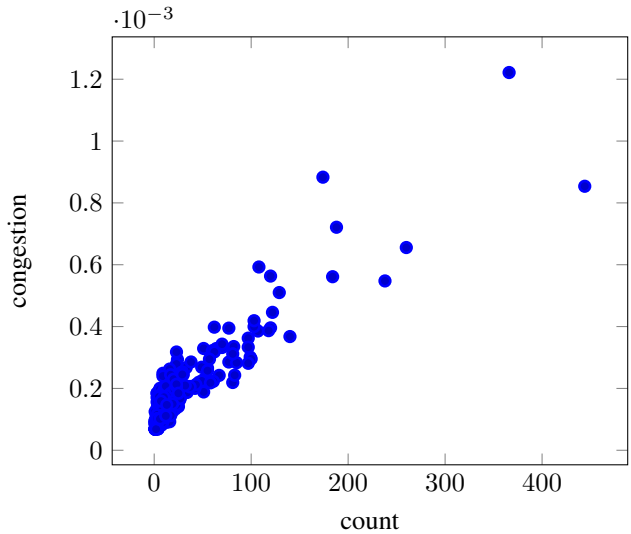

The above definition of congestion would be the same for every image where there is no overlap. However, when there are many overlaps, the number of nonzero pixels should be lower. Thus, $C$ will be larger. After the test set is divided, we compare the subset average results. In Figure \ref{fig:countcongcorr}, we see the correlation between the ground-truth count and the defined congestion for the TH2019 dataset.

\begin{figure}[b!]
  \begin{center}
    \includegraphics[width=\linewidth]{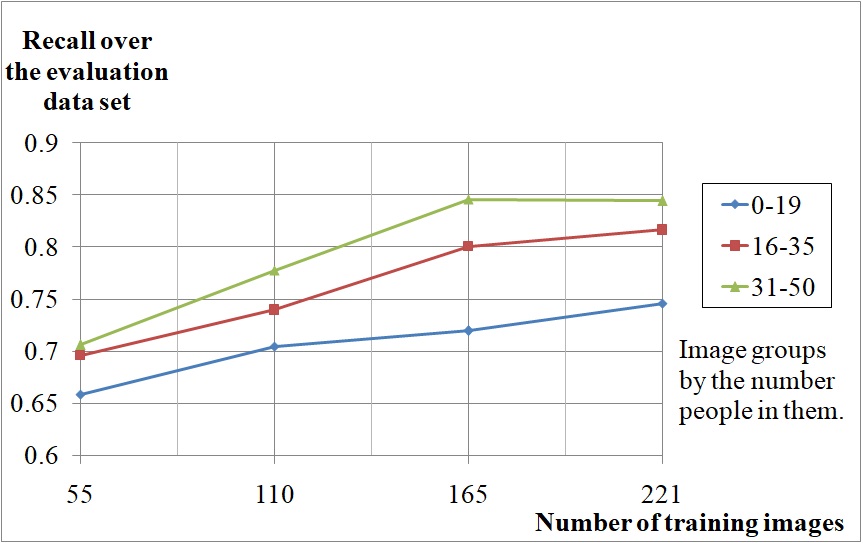}
    \caption{Average recall ratio of Faster R-CNN network using different amounts of training images}
    \label{fig:frcnnrec}
  \end{center}
\end{figure}

\subsubsection{Count-based congestion}
In the other approach, the congestion level is the count. We split the complete dataset into smaller sub-datasets by their ground-truth count. We needed a method to decide the partition sizes. The dataset was already small, and with further reduction, we risked overfitting. Prior to analysis, we had a preliminary research. In that we used a smaller dataset and only experimented with Faster R-CNN. We investigated the recall with regards to different training-set sizes and pedestrian counts. The results can be seen in Figure \ref{fig:frcnnrec}. From this, we can see that the number of pedestrians matter more than the number of training images, because images having more people have higher recall. Therefore, the partitions should have roughly the same total pedestrian counts. Thus, the tests can be less affected by the differences between datasets.

All subsets contained roughly 3,000 annotations. Thus, there were overlaps. We put the images in increasing order by pedestrian count. The low congestion group comprised images of the first 3,000 pedestrians. The high congestion group contained the last 3,000 pedestrians, and the medium congestion group contained between 1,500 pedestrians before and 1,500 after the mid-most pedestrian. Of course, we did not begin to split images at that point. We instead separated the subsets after the first image went past the required pedestrian count. Thus, the low, medium, and high partitions contained 253, 43, and 17 images, respectively. As we can see, the image count was rather low in the medium- and high-congestion sets.

The subsets were divided into training and test sets the same way as the complete set. We trained one model of each network for each subset and compared them.

\subsection{Experiments}

In this section, we discuss the specific networks we trained. For every method, four models were trained: one on the complete set and one for each subset. We attempted using pre-trained weights trained on MSCOCO and Shanghaitech, but it did not really make a difference. Thus, we trained from scratch.

\begin{itemize}
\item{Object detection methods}
\begin{itemize}
\item{Faster R-CNN Inception Resnet v2 Atrous}
\item{YOLOv3 (original)}
\item{YOLOv3 (one detector)}

This network was our own creation, although it is only a slightly modified version of the original YOLOv3 model. We removed the two larger-scale detectors, which means six layers. Additionally, we increased the number of priors-per-cell to five and adjusted the prior sizes to better fit our data. Thus, we optimized the loss function to our problem.
\end{itemize}

\item{Density map methods}
\begin{itemize}
\item{MCNN}
\item{CSRNet}
\item{CAN}
\end{itemize}
\end{itemize}

\begin{figure*}[t!]
\centering
\begin{subfigure}{.19\textwidth}
  \centering
  \includegraphics[width=\textwidth]{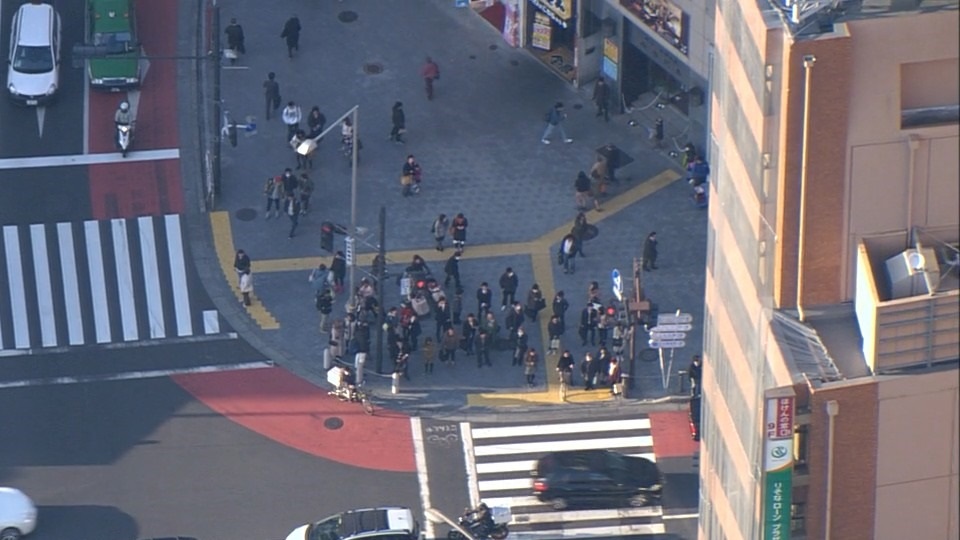}
  \captionsetup{belowskip=0.3cm}
  \caption{Original image}  
\end{subfigure}
\begin{subfigure}{.19\textwidth}
  \centering
  \includegraphics[width=\textwidth]{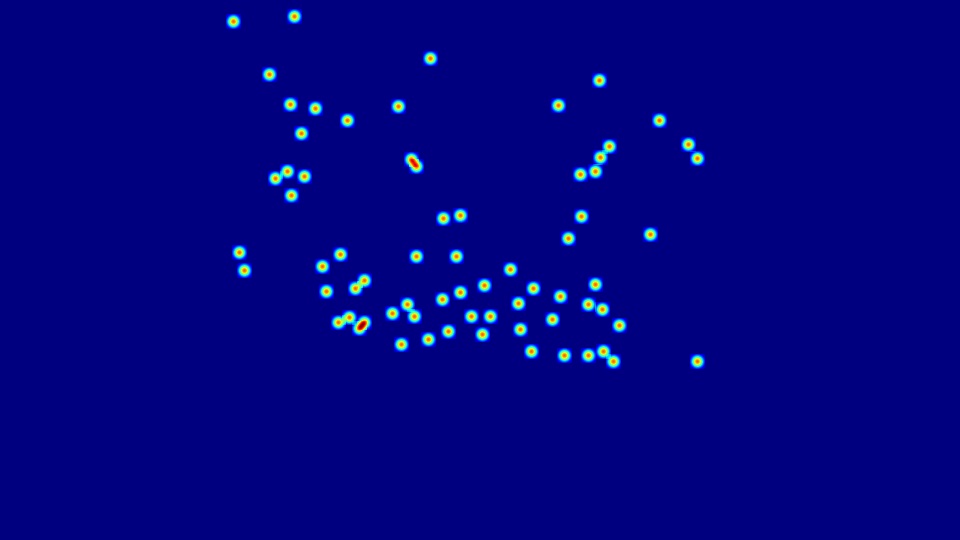}
  \captionsetup{belowskip=0.3cm}
  \caption{Rough ground-truth}
\end{subfigure}
\begin{subfigure}{.19\textwidth}
  \centering
  \includegraphics[width=\textwidth]{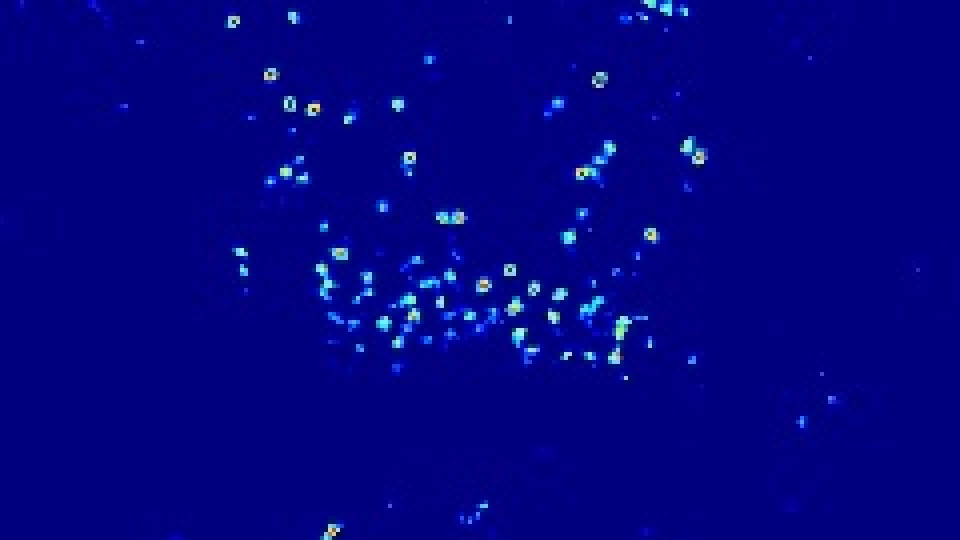}
  \captionsetup{belowskip=0.3cm}
  \caption{MCNN}
\end{subfigure}
\begin{subfigure}{.19\textwidth}
  \centering
  \includegraphics[width=\textwidth]{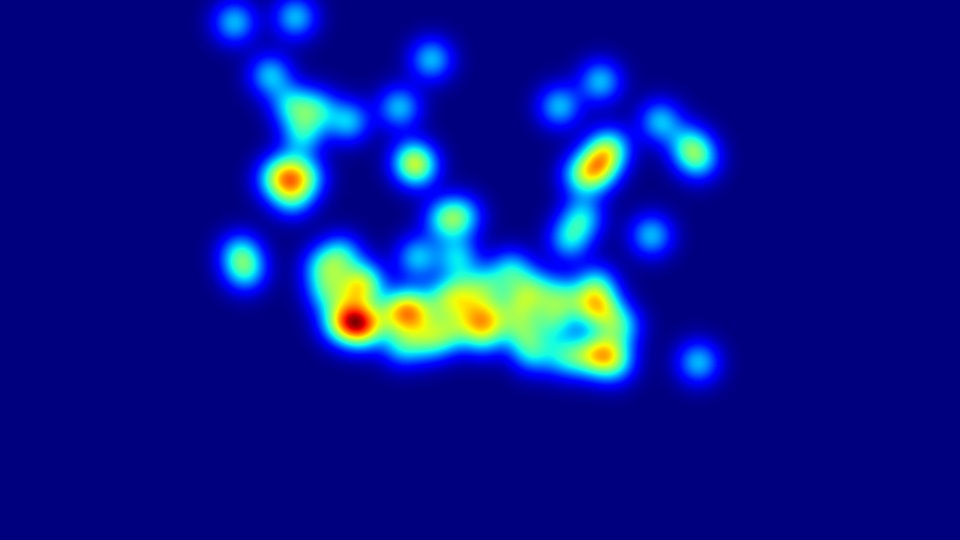}
  \captionsetup{belowskip=0.3cm}
  \caption{Smooth ground-truth}
\end{subfigure}
\begin{subfigure}{.19\textwidth}
  \centering
  \includegraphics[width=\textwidth]{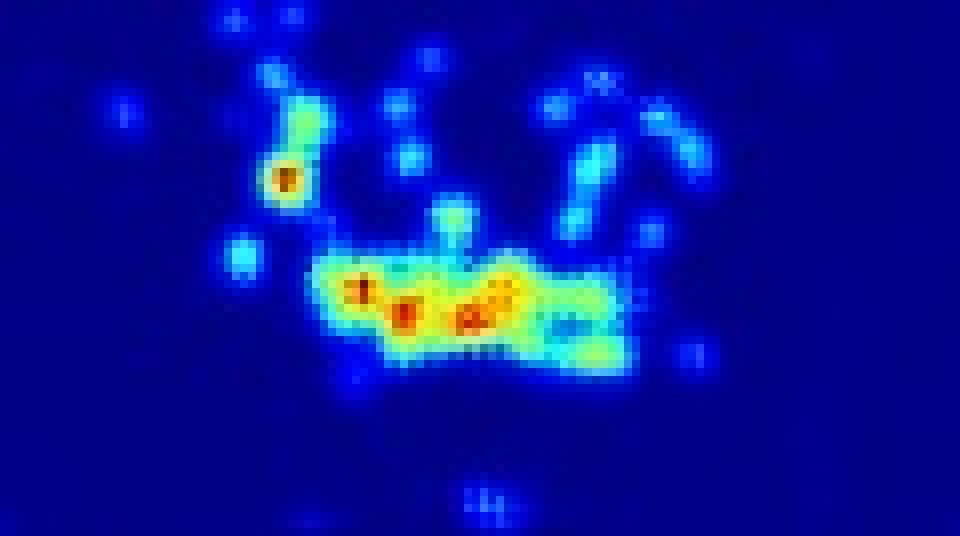}
  \captionsetup{belowskip=0.3cm}
  \caption{CAN}
\end{subfigure}
\begin{subfigure}{.19\textwidth}
  \centering
  \includegraphics[width=\textwidth]{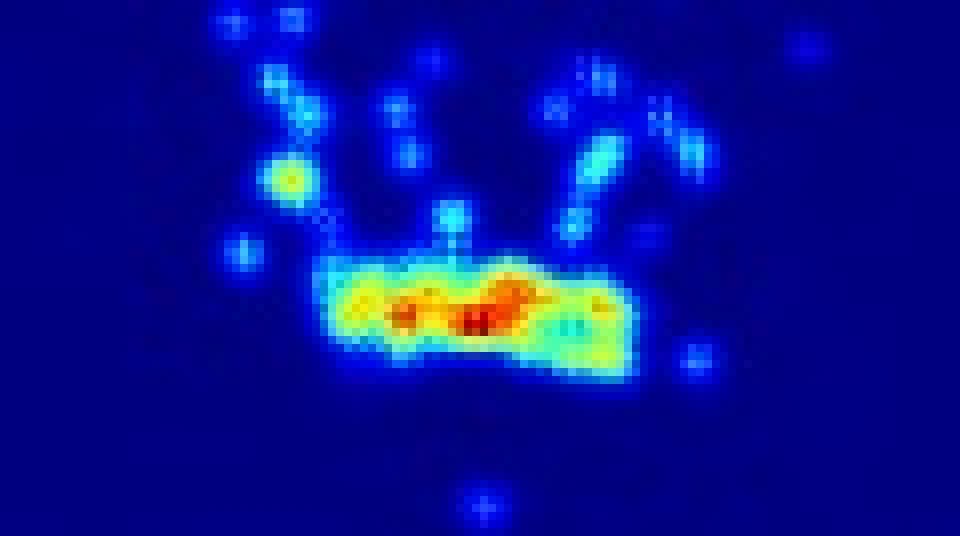}
  \captionsetup{belowskip=0.3cm}
  \caption{CSRNet}
\end{subfigure}
\begin{subfigure}{.19\textwidth}
  \centering
  \includegraphics[width=\textwidth]{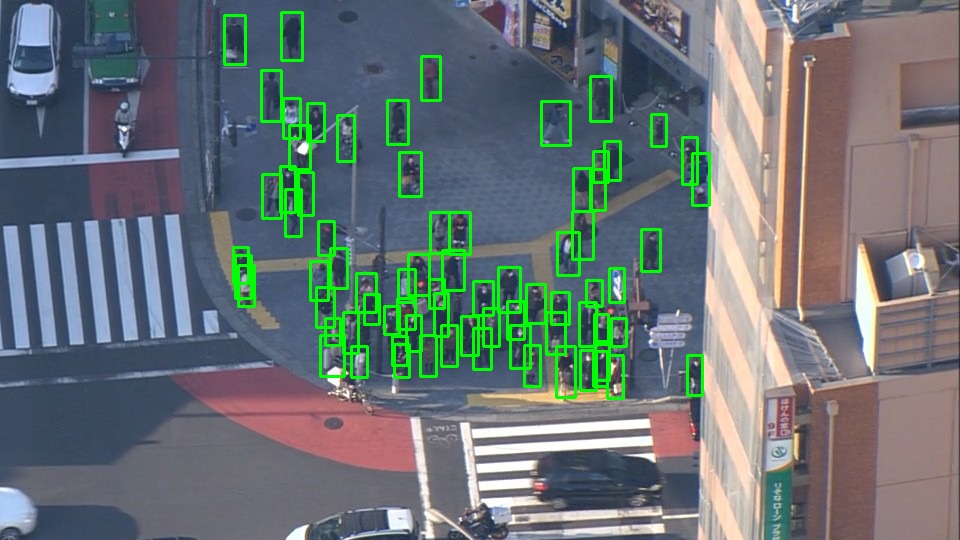}
  \captionsetup{belowskip=0.3cm}
  \caption{Faster R-CNN}
\end{subfigure}
\begin{subfigure}{.19\textwidth}
  \centering
  \includegraphics[width=\textwidth]{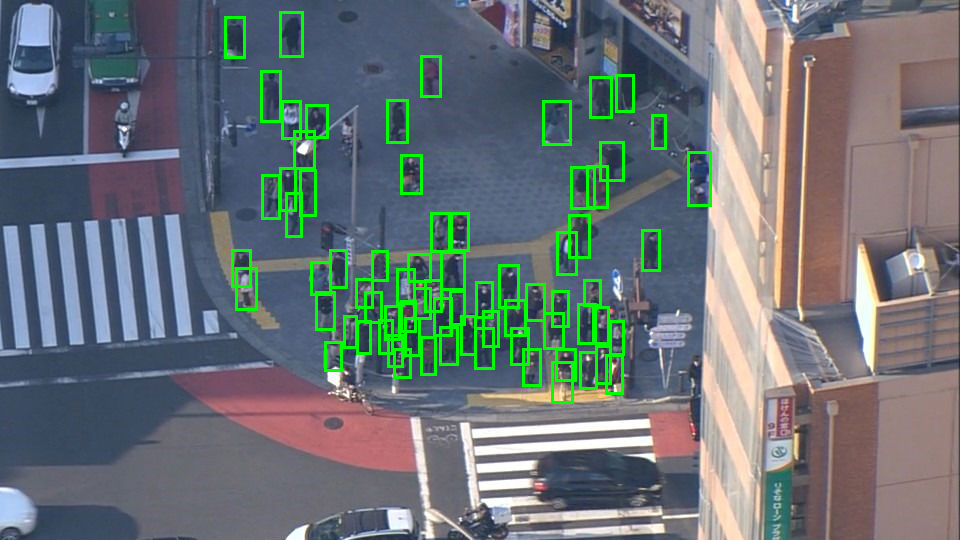}
  \captionsetup{belowskip=0.3cm}
  \caption{YOLO (one detector)}
\end{subfigure}
\begin{subfigure}{.19\textwidth}
  \centering
  \includegraphics[width=\textwidth]{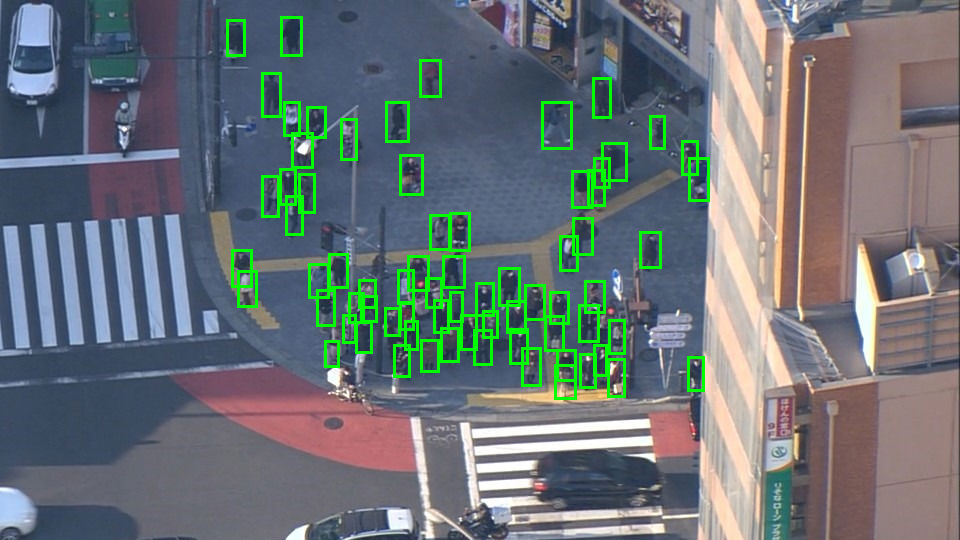}
  \captionsetup{belowskip=0.3cm}
  \caption{YOLO (original)}
\end{subfigure}
\caption{Ground-truth and estimation results for networks trained on the complete TH2019 dataset.}
\label{fig:detectionresults}
\end{figure*}

\section{Results and Discussion}\label{Results and Discussion}

\begin{table}[b!]
  \begin{center}
      \begin{filecontents*}{CSV/overallresults.csv}
Method,Precision,Recall,MAE,RMSE
Faster R-CNN Inception Resnet v2 Atrous,0.8813,0.7269,6.33,13.85
YOLOv3 (original),0.9,0.72,6.25,13.76
YOLOv3 1 detector (our),0.89,0.76,4.85,9.46
MCNN,,,13.28,16.65
CSRNet,,,3.82,6.36
CAN,,,3.17,4.39
\end{filecontents*}
\pgfplotstabletypeset[
      font=\footnotesize,
      col sep=comma,
      header=has colnames,
      display columns/0/.style={column type={p{29mm}},string type},
      every head row/.style={before row=\toprule,after row=\bottomrule},
      every last row/.style={after row=\bottomrule},
      every even row/.style={before row=\rowcolor[gray]{0.9}},
      every row 1 column 1/.style={postproc cell content/.style={@cell content=\bf{\num{##1}}}},
      every row 2 column 2/.style={postproc cell content/.style={@cell content=\bf\num{##1}}},
      every row 5 column 3/.style={postproc cell content/.style={@cell content=\bf\num{##1}}},
      every row 5 column 4/.style={postproc cell content/.style={@cell content=\bf\num{##1}}},
      empty cells with={$-$},
      assign column name/.style={/pgfplots/table/column name={\textbf{#1}}},
    ]{CSV/overallresults.csv}
    \caption{Overall results for the tested networks over the complete TH2019 dataset. Recall and precision are reported for IoU=0.5. Best values are in boldface.}
    \label{Tab:overallres}
  \end{center}
\end{table}

\subsection{Overall accuracy}
In Table \ref{Tab:overallres}, the different metrics for all trained networks are shown for the complete TH2019 evaluation dataset. The most important information found here pertains to the density map-based methods, which dominate. However, the MCNN does not perform so great. The CAN method yields the best counts. Therefore, it is the most valuable prospect for our target application. We see a general difference between the object detection-and density map-based methods. With the density map-based methods, the MAE and RMSE values were much closer. Thus, the variation of errors was much smaller from image to image. Therefore, the prediction was more reliable. Faster RCNN and the original YOLOv3 were approximate, whereas our modified YOLOv3 network outperformed them significantly in recall, MAE, and RMSE. However, it was really close in precision to YOLOv3. Figure \ref{fig:detectionresults} illustrates the results.

If we further investigate CAN, we would make more observations. In some situations, where a billboard contains images of people, the network detects these images as pedestrians. Furthermore, bicycle and motorbike riders are also often detected. Thus, the model appears to be underfitted. In Table \ref{Tab:cancomp}, we compare the results of CAN on the TH2019 dataset with other benchmark datasets. The MAE and RMSE values correlated with the average count. Thus, we presented the relative values. From the other datasets, we see that the relative error increased as the average count increased. However, the TH2019 dataset did not fit this trend. The average count was much less, whereas the relative error was between the values of the other sets. The very-small target sizes and the low amount of training data might have caused this. Regarding the underfitting, there was not much we could do to address this.

\begin{table}[h!]
  \begin{center}
      \resizebox{\linewidth}{!}{
      \begin{filecontents*}{CSV/comparedatasets.csv}
Dataset,MAE,MSE,Average count,rel.MAE,rel.MSE
TH2019,3.17,4.39,30,0.105666667,0.146333333
ShanghaiTechB,7.8,12.2,123,0.063414634,0.099186992
ShanghaiTechA,62.3,100,501,0.124351297,0.199600798
UCF\_CC\_50,212.2,243.7,1279,0.165910868,0.190539484
\end{filecontents*}
\pgfplotstabletypeset[
      col sep=comma,
      header=has colnames,
      display columns/0/.style={column type={l},string type},
      display columns/1/.style={column type={S},string type},
      display columns/2/.style={column type={S},string type},
      display columns/3/.style={column type={S},string type},
      display columns/4/.style={column type={S},string type},
      display columns/5/.style={column type={S},string type},
      every head row/.style={before row=\toprule,after row=\bottomrule},
      every last row/.style={after row=\bottomrule},
      every even row/.style={before row=\rowcolor[gray]{0.9}},
      empty cells with={$-$},
      assign column name/.style={/pgfplots/table/column name={\textbf{#1}}},
    ]{CSV/comparedatasets.csv}}
    \caption{Comparison of CAN results among various datasets.}
    \label{Tab:cancomp}
  \end{center}
\end{table}

\subsection{Cross-validation}
To show that our dataset provides contribution we evaluate it on models trained from other datasets. Table \ref{Tab:crossval} shows the accuracy of our selected architectures on the TH2019 complete validation set with weights trained from different datasets. All models that were trained on TH2019 perform significantly better than their non-TH2019 counterparts.

\begin{table}[t!]
  \begin{center}
      \resizebox{\linewidth}{!}{
      \begin{filecontents*}{CSV/crossval.csv}
Method and dataset, mAP @IoU=0.5, relative MAE
Faster RCNN Inception Resnet v2 Atrous + MSCOCO, 20.08{\%},
Faster RCNN Inception Resnet v2 Atrous + TH2019, 67.00{\%},
YOLOv3 + MSCOCO, 11.80{\%},
YOLOv3 + VisDrone2019, 36.25{\%},
YOLOv3 + TH2019, 81.47{\%},
YOLOv3  1 detector + TH2019, 82.91{\%},
MCNN + ShanghaiTech B,,1.09
MCNN + TH2019,,0.44
CSRNet + ShanghaiTech B,,1.08
CSRNet + TH2019,,0.13
CAN + ShanghaiTech A,,2.35
CAN + TH2019,,0.1
\end{filecontents*}
\pgfplotstabletypeset[
      col sep=comma,
      header=has colnames,
      string type,
      display columns/0/.style={column type={p{5cm}}},
      display columns/1/.style={column type={p{1.5cm}}},
      display columns/2/.style={column type={p{1.2cm}}},
      every head row/.style={before row=\toprule,after row=\bottomrule},
      every last row/.style={after row=\bottomrule},
      every even row/.style={before row=\rowcolor[gray]{0.9}},
      empty cells with={$-$},
      assign column name/.style={/pgfplots/table/column name={\textbf{#1}}},
    ]{CSV/crossval.csv}}
    \caption{Comparison between models trained on other dataset and models trained on TH2019  when evaluated on TH2019 (we use only one class).}
    \label{Tab:crossval}
  \end{center}
\end{table}

Images in the VisDrone2019 dataset have the most common attributes with ours. Even the model trained on this dataset performs significantly worse on the TH2019 validation dataset than our YOLOv3 1 detector model trained on TH2019 training set. Based on these findings we can say that our dataset - as small as it might be - delivers features which are not covered by other pedestrian datasets.

\begin{figure}[t!]
\centering
\begin{tikzpicture}
  \begin{axis}[width=8cm,height=4cm,symbolic x coords={Low,Medium,High},xlabel=Count based congestion,ylabel=$\frac{MAE}{AVG Count}$,xtick=data,legend style={at={(1,1.1)},anchor=south east,legend columns={3}},
                       legend entries={M-CNN,CSRNet,CAN,Faster R-CNN,YOLOv3 3D,YOLOv3 1D},ymin=0,ymax=1]
    \addplot[red,mark=star] table [x=Method, y=M-CNN, col sep=comma] {CSV/maepercent.csv};
    \addplot[green,mark=triangle*] table [x=Method, y=CSRNet, col sep=comma] {CSV/maepercent.csv};
    \addplot[gray,mark=square*] table [x=Method, y=CAN, col sep=comma] {CSV/maepercent.csv};
    \addplot[blue,mark=diamond*] table [x=Method, y=Faster R-CNN, col sep=comma] {CSV/maepercent.csv};
    \addplot[black] table [x=Method, y=YOLOv3 3D, col sep=comma] {CSV/maepercent.csv};
    \addplot[yellow,mark=o] table [x=Method, y=YOLOv3 1D, col sep=comma] {CSV/maepercent.csv};
  \end{axis}
\end{tikzpicture}
\caption{MAE values for different count levels relative to the average people-count per image.}
\label{fig:cong1}
\end{figure}
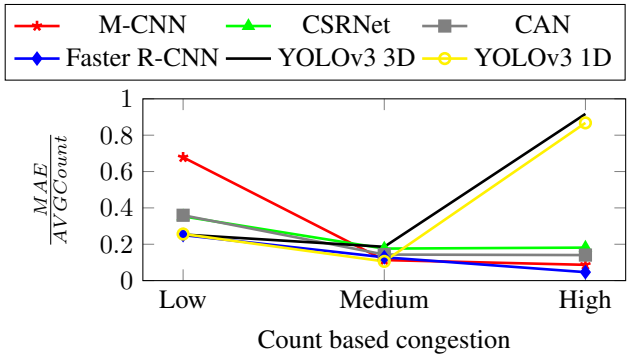

\subsection{Count-based congestion analysis}

Figure \ref{fig:cong1} shows the results of congestion analysis performed by training the methods using datasets of different average counts. Most methods showed increasing accuracy as the average count grew. In the analysis, YOLO exhibited a different behavior. With highly congested images, the accuracy dropped significantly. This suggests that single-stage detectors do not perform well when trained on congested images. Another possibility is that the number of training images was too low. With Faster R-CNN, the classifier is trained on every region proposal. The density map-based methods were trained on several hundred smaller image patches. YOLO was trained on whole images end-to-end with one passthrough, which involved far less training data for classifiers. Nevertheless, the single-stage detector had considerably worse accuracy than the other methods when it was trained on the high congested subdataset.


\subsection{Density-based congestion analysis}
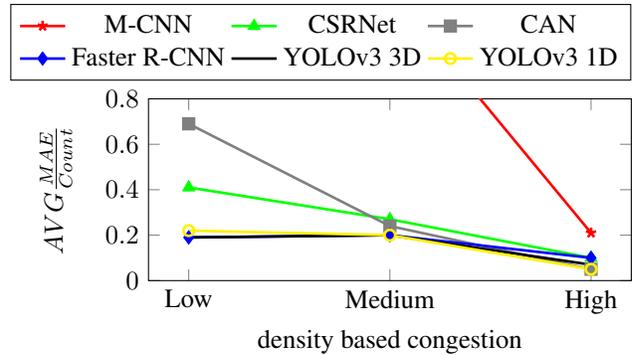
\begin{figure}[t!]
\centering
\begin{tikzpicture}
  \begin{axis}[width=8cm,height=4cm,symbolic x coords={Low,Medium,High},xlabel=density based congestion,ylabel=$AVG\frac{MAE}{Count}$,xtick=data,legend style={at={(1,1.1)},anchor=south east,legend columns={3}},
                       legend entries={M-CNN,CSRNet,CAN,Faster R-CNN,YOLOv3 3D,YOLOv3 1D},ymin=0,ymax=0.8]
    \addplot[red,mark=star] table [x=Method, y=M-CNN, col sep=comma] {CSV/congmaepercent.csv};
    \addplot[green,mark=triangle*] table [x=Method, y=CSRNet, col sep=comma] {CSV/congmaepercent.csv};
    \addplot[gray,mark=square*] table [x=Method, y=CAN, col sep=comma] {CSV/congmaepercent.csv};
    \addplot[blue,mark=diamond*] table [x=Method, y=Faster R-CNN, col sep=comma] {CSV/congmaepercent.csv};
    \addplot[black] table [x=Method, y=YOLOv3 3D, col sep=comma] {CSV/congmaepercent.csv};
    \addplot[yellow,mark=o] table [x=Method, y=YOLOv3 1D, col sep=comma] {CSV/congmaepercent.csv};
  \end{axis}
\end{tikzpicture}
\caption{MAE values for different congestion levels relative to the average people-count per image}
\label{fig:cong2}
\end{figure}

In this section, we analyze the congestion against the complete evaluation dataset split into three subsets. Because the images in the same group had vastly different annotation counts, we took the MAE values relative to the individual counts and averaged them. The fundamental difference between this method and the count-based one entails this case having a model that was trained on a wide range of data. The results shown in Figure \ref{fig:cong2} confirm our previous findings. The object-detection-based solutions show stagnation between low and medium congestion. However, the rest shows clear improvements of accuracy with an increase of congestion level. M-CNN accuracy at lower congestion levels is very bad. Thus, if were to plot it, the rest of the diagram would be unreadable.

When examining the results skeptically, the most obvious explanation for this behavior is that the training methods overfitted the scenes in the congested situation. Although the training images and the evaluation images were completely different, we had a very limited number of scenes, especially congested ones. Doing a random split on a dataset with a lot more images than scenes inevitably causes the training and test sets both to contain images of the same scene. The only way to confirm this was to increase the size of the dataset. Training and test datasets could be then split into subsets containing different scenes.

Another less skeptical explanation is one that assumes the methods had some kind of offset-like error. For the density map-based methods, empty areas were not actually empty. Instead, they had a rather low density value. As congestion increased, the size of these areas decreased. Furthermore, the ground-truth count was larger. Thus, proportionally the error was less, which better explains the accuracy. We cannot contemplate similar errors that are simple to understand in the case object detection networks. Our theory is that if the feature extractor were to identify something as large as a clump of people and recognize its size, it could just randomly drop boxes on it. If it were to drop enough boxes, eventually all targets would be identified, and false detections would be eliminated by non-maximum suppression. Because these networks make several thousands or tens-of-thousands of predictions, this is an entirely plausible scenario that is backed by our findings.

These findings mean that a network architecture, which combines other models and switches between them based on their congestion level, would not bring any significant change to their accuracy.

\section{Conclusions}

In this paper we surveyed a number of CNN-based methods for the task of pedestrian counting using a custom aerial footage. The methods comprised two types: density map- and object detection-based solutions. For the investigation, we introduced a dataset. During our investigation, we used several open-source implementations for the selected methods. Doing so, we arrived at the conclusion that the investigated density map-based solutions performed significantly better than the object detection-based ones. However, we found negative examples among them.

The introduced dataset exhibits high altitude, steep angle viewpoint, and low perspective distortion. It comprises crowded urban images. Our dataset cross-validation results show that pedestrian detection on this kind of data is not covered by other datasets.

We also successfully optimized the loss function of a general-purpose object detection method to better fit the introduced dataset. Thus, we significantly increased accuracy. We exploited the fact that we had preliminary knowledge about the dataset and the target sizes.

We also conducted a congestion analysis on the methods to see how they performed with different crowd levels. Our findings show that, as the congestion increased, the accuracy stagnated or got better for most methods, regardless of type. Our investigations were conducted for real-life situations, and, from a practical perspective, the above observations were well-made.
{\small
\bibliographystyle{ieee_fullname}
\bibliography{refs}
}

\end{document}